\documentclass[letterpaper, 10 pt, journal, twoside]{IEEEtran}
\usepackage{subfig,float}
\usepackage{cite}
\usepackage{amsmath,amssymb,amsfonts}
\usepackage{algorithmic}
\usepackage{graphicx}
\usepackage{booktabs}

\usepackage{textcomp}
\usepackage{xcolor}
\def\BibTeX{{\rm B\kern-.05em{\sc i\kern-.025em b}\kern-.08em
    T\kern-.1667em\lower.7ex\hbox{E}\kern-.125emX}}

\usepackage{mathtools}
\usepackage{algorithm}
\usepackage{comment}
\usepackage{mathbbol}
\usepackage{amsmath,amsfonts,amssymb}

\newcommand{\X}{\mathbb{X}}

\newcommand{\U}{\mathbb{U}}

\newcommand{\V}{\mathbb{V}}

\newcommand{\R}{\mathbb{R}}

\newcommand{\EX}{\mathfrak{X}}
\newcommand{\ex}{\mathfrak{x}}
    
\newtheorem{problem}{Problem}
\newtheorem{remark}{Remark}
\newtheorem{assumption}{Assumption}

\begin{document}
%
\title{Evolutionary Gait Reconfiguration in Damaged Legged Robots}
%
%
%


\author{Sahand Farghdani and Robin Chhabra, ~\IEEEmembership{Senior Member,~IEEE}%
\thanks{S. Farghdani is with the Mechanical and Aerospace Engineering
Department, Carleton University, Ottawa, ON K1S 5B6, Canada.{\tt\small sahandfarghdani@cmail.carleton.ca}
        \\
\indent Robin Chhabra (corresponding author) is with the Mechanical, Industrial, and Mechatronics Engineering
Department, Toronto Metropolitan University, Toronto, ON M5B 2K3, Canada.
{\tt\small robin.chhabra@torontomu.ca}}}%

%
%

\markboth{IEEE Robotics and Automation Letters.}
{
Damage recovery in legged robots} 

%



\maketitle

\begin{abstract}

Multi-legged robots deployed in complex missions are susceptible to physical damage in their legs, impairing task performance and potentially compromising mission success. This letter presents a rapid, training-free damage recovery algorithm for legged robots subject to partial or complete loss of functional legs. The proposed method first stabilizes locomotion by generating a new gait sequence and subsequently optimally reconfigures leg gaits via a developed differential evolution algorithm to maximize forward progression while minimizing body rotation and lateral drift. The algorithm successfully restores locomotion in a 24-degree-of-freedom hexapod within one hour, demonstrating both high efficiency and robustness to structural damage.

\end{abstract}


\IEEEpeerreviewmaketitle

\section{Introduction}

\IEEEPARstart{D}{amage} recovery capabilities have become essential as robots are increasingly deployed beyond controlled laboratory environments. In Multi-Legged Robots (MLRs), partial or complete loss of leg functionality alters morphology, compromising locomotion and stability. One of the most immediate consequences is falling, which has been extensively addressed in recent works \cite{sun2024fall,li2024dynamic,lu2024learning,ma2023learning}.

Early damage recovery strategies primarily relied on external monitoring systems to trigger precomputed responses—such as gait adaptation in hexapods with impaired limbs \cite{zhang2024gait}. However, these approaches were vulnerable to communication delays and failures. More recent efforts emphasize autonomous self-diagnosis and recovery, though these often require complex sensing and high computational demands, limiting their applicability in time-critical missions such as search and rescue.
Autonomous recovery methods can generally be classified into three categories: healing, replacement, and reconfiguration. Healing-based strategies exploit self-healing materials \cite{ref20}, including photo-weldable shape memory composites \cite{ref21} and neural cellular automata capable of regenerating damaged morphology \cite{ref22}. In rigid robots, replacement-based fault tolerance is more common. For example, \cite{ref26} presents a multi-robot system in which a secondary robot replaces a failed unit, while \cite{ozkan2021self} describes neighboring agents transporting a damaged robot to a repair site.

Reconfiguration approaches, initially explored in soft robotics, are increasingly applied to rigid robots with redundant Degrees of Freedom (DoF) through adaptive control and path planning. Bongard’s seminal work introduced a self-modeling robot that inferred its structure via actuation-sensation feedback to recover locomotion \cite{ref14}. Building on this, \cite{ref17,cully2015,ref18} proposed an intelligent trial-and-error method in which a robot generates a performance map in simulation and selectively tests high-potential gaits in reality, achieving rapid adaptation. 
Control strategies have also been developed to facilitate damage recovery. Farid \textit{et al.} implemented sliding mode control to mitigate actuator faults such as bias, gain degradation, and saturation \cite{farid2018fractional}. Suzuki \textit{et al.} proposed a hybrid controller that switches between model-based and learning-based strategies to compensate for unanticipated behaviors \cite{ref27}. Feber \textit{et al.} introduced dynamic adaptation rules for rhythmic gait re-learning following leg damage \cite{feber2022gait}, while Yang \textit{et al.} demonstrated compensatory gait generation for quadrupeds with simulated joint failure \cite{yang2002fault}. Johnson \textit{et al.} applied gait classification for hexapods with single-actuator legs to restore stable walking after leg failure \cite{johnson2010disturbance}.

The majority of existing work focuses on modifying gait patterns to accommodate morphological changes. Accordingly, robust gait generation strategies are critical to ensure post-damage functionality. Christensen \textit{et al.} presented a fault-tolerant gait learning framework using stochastic optimization of central pattern generator parameters, achieving convergence within ten minutes post-failure \cite{christensen2014fault}. Pratihar \textit{et al.} developed a Genetic Algorithm (GA)-fuzzy method to jointly optimize gait trajectories and sequences in six-legged robots \cite{pratihar2002optimal}. Erden \textit{et al.} used reinforcement learning to achieve adaptive free gait generation following rear-leg failure in a hexapod \cite{erden2008free}, while Shi \textit{et al.} extended this to quadrupeds using evolutionary radial basis function networks for trajectory adaptation across diverse terrains \cite{shi2022reinforcement}. Zhang \textit{et al.} analyzed gait duty factors for minimizing foot slip and attitude fluctuation in hexapods \cite{zhang2021straight}, and Kon \textit{et al.} explored GA-based gait generation for damaged robots, though real-world applicability remained limited due to gait irregularity \cite{kon2020gait}. These concepts have also been extended to wheeled-legged systems \cite{bjelonic2021whole}.

This letter presents a novel damage recovery algorithm that first stabilizes an MLR's motion by generating an appropriate gait sequence for the remaining functional legs, followed by trajectory optimization for each leg using a Differential Evolution (DE) algorithm. The objective is to maximize forward locomotion while minimizing body oscillations and lateral drift. The DE optimization is grounded in a validated modular dynamic modeling framework \cite{future:farghdani_model2}, which enables accurate, faster-than-real-time simulation of damaged robot dynamics through the adoption of a morphology vector. We validate the proposed framework through extensive experiments on a 24-DoF hexapod robot, demonstrating its computational efficiency and effectiveness across a range of damage scenarios, including single and dual leg loss. Notably, the robot’s locomotion is reliably restored in under one hour without the need for prior training or behavior-performance mapping.
The key contributions of this work are:

\begin{enumerate}
\item Unlike prior methods \cite{feber2022gait, ozkan2021self, ref26}, our algorithm enables accurate damage recovery without relying on complex sensor arrays or external monitoring.

\item In contrast to \cite{ref17, cully2015, ref18, ref27}, we explicitly leverage validated closed-form dynamics of the damaged robot for gait reconfiguration, avoiding the need for initial training or trial-and-error testing.

\item Unlike most existing methods \cite{johnson2010disturbance, zhang2024gait, erden2008free}, our approach can recover from multiple leg failures.
\end{enumerate}

This letter is organized as follows: Section \ref{sec:prob-stat} states the problem tackled in this letter. In Section \ref{chap6:method}, we present our damage recovery algorithm. We investigate the accuracy of our algorithm in multiple case studies, in Section \ref{chap6:results}. Section \ref{chap6:conclusion} includes some concluding remarks.

\section{Problem Statement}\label{sec:prob-stat}

The healthy MLR considered in this letter consists of a main body indexed by $b$, $N$ legs indexed by $i\in\{1, \,...\, ,N\}$, and the $i^{th}$ leg includes $n_i$ links indexed by $j\in\{1, \,...\, ,n_i\}$, so $l_{ij}$ represents Link $j$ of Leg $i$. The total number of robot's leg DoF is defined by $N_T:=\sum_{i=1}^{N}n_i$; hence, the system has $6+N_T$ degrees of freedom. 
When an MLR experiences physical damage, its morphology changes due to lost or malfunctioning limbs.  
To represent the post-damage morphology, we define the leg morphology vector:
    \begin{align} \label{chap6:Ex}
    &\EX := \begin{bmatrix} \ex_1 & \cdots & \ex_i & \cdots & \ex_N      
    \end{bmatrix}\in\R^{N},  \\
    &\ex_{i}=\left\{ \begin{array}{rcl}
     1 & \mbox{if Leg $i$ is functional} \\ 
     0 & \mbox{otherwise} 
    \end{array}\right..
\end{align}
The leg morphology vector is a binary row vector of length $N$, where each element corresponds to one of the robot’s legs. In the undamaged case, all entries of $\EX$ are set to one. When a leg becomes nonfunctional, the corresponding entry is set to zero, effectively encoding the robot's structural state. Before formally stating the problem, we outline the key assumptions adopted in this work.

\begin{assumption}\label{chap6:as1}
    We assume having access to the damaged robot’s morphology vector $\EX$ and the nominal gait designed for the healthy system.
\end{assumption}

\begin{assumption}\label{chap6:as2}
   A leg is considered nonfunctional even if only certain links or joints within it are impaired. For the purpose of this study, we assume that nonfunctional legs are completely detached from the robot, thereby contributing neither to locomotion nor stability. In realistic scenarios such as partial structural damage or joint locking we assume that the robot is capable of repositioning the impaired leg(s) to avoid obstructing the motion of the functional limbs.
\end{assumption}

\begin{problem}\label{chap6:pr1}
Given a unique leg morphology vector $\EX$ for a damaged MLR, and under Assumptions \ref{chap6:as1} and \ref{chap6:as2}, determine an optimal gait reconfiguration to restore forward locomotion in the robot. This includes identifying a new gait sequence and an optimal set of tip-end trajectories for the remaining functional legs, such that forward motion is maximized while minimizing body oscillations and lateral deviation from a straight path.
\end{problem}


\section{Methodology}\label{chap6:method}

This section presents the proposed damage recovery framework for solving Problem \ref{chap6:pr1}, which is based on gait reconfiguration of the damaged MLR. The gait of an MLR consists of two fundamental components: the gait sequence and the leg tip trajectories. The leg tip trajectory governs the smooth forward progression of the robot’s main body, while the gait sequence ensures stability by defining the activation order of legs and distinguishing between support and swing phases at any given time. These components are inherently coupled through the phase differences between leg trajectories. The framework jointly optimizes both components to achieve stable and effective locomotion during recovery. 

To minimize the risk of further damage, the framework eliminates the need for trial-and-error deployment of candidate gaits on the physical robot—an approach commonly adopted in existing recovery strategies. Instead, the method performs an entirely offline optimization, utilizing a high-fidelity dynamic model of the damaged system. The optimization engine is based on a DE algorithm, which searches for a solution that maximizes forward locomotion while minimizing body oscillations and lateral drift. The cost function is evaluated via a fast, modular simulation engine specifically developed for modeling the dynamics of damaged MLRs, as detailed in \cite{farghdani2024singularity, future:farghdani_model2}. This faster-than-real-time simulator employs a Boltzmann-Hamel Lagrangian formulation to accurately capture the behavior of the robot under various damage scenarios.


   


\subsection{Gait Sequence Planning}

When an MLR experiences structural damage, maintaining the stability of its main body becomes the primary challenge. Stability requires that the robot’s Center of Mass (CoM) remains within the support polygon defined by its legs in contact with the ground. Typically, at least three legs must be in the support phase at all times. Consequently, to enable effective recovery and sustained locomotion, the robot must retain at least two functional legs on each side. 

The initial step in the proposed recovery strategy is the design of a new gait sequence that prioritizes body stability under the current morphological constraints of the robot. For a hexapod robot, there exist three canonical gait sequences that satisfy the criterion of maintaining at least three supporting legs throughout locomotion. These sequences serve as the basis for selecting a feasible and stable recovery gait:
\begin{itemize}
    \item \textbf{Tripod gait}: Divides the legs into two alternating groups of three legs, providing three-leg support at all times.  
    \item \textbf{Quadrangular gait}: Divides the legs into three alternating groups of two legs, providing four-leg support at all times.  
    \item \textbf{Pentagonal gait}: Ensures continuous five-leg support during the gait cycle with only one leg entering swing at a time.  
\end{itemize}
\begin{remark}
Without loss of generality, the following discussion is limited to canonical gaits for hexapod robots. With appropriate modifications, these gait patterns can be extended to MLRs with more or fewer than six legs, provided that the stability condition is satisfied.\end{remark}

\begin{figure}[hbtp]
  \centering
  \includegraphics[width=.35\textwidth]{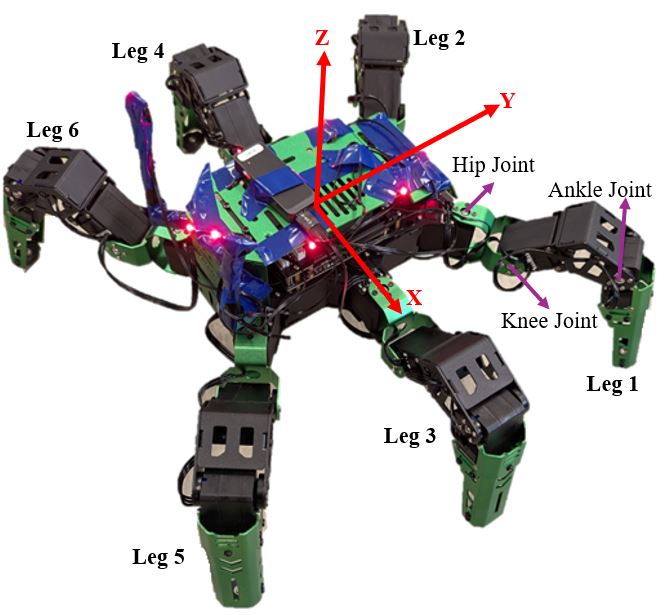}\\
  \caption{Hiwonder JetHexa robot }
  \label{chap6:Robot}
\end{figure}

In our gait sequence planning, a healthy hexapod robot always employs the tripod gait as its default locomotion strategy. In this letter, the nominal gait refers specifically to a tripod pattern in which Legs with index in the set $\mathcal L_1=\lbrace 1, 4, 5 \rbrace$ operate in synchrony, forming the first leg group, while Legs indexed by $\mathcal L_2=\lbrace 2, 3, 6 \rbrace$ comprise the second group. The leg numbering convention is illustrated in Fig. \ref{chap6:Robot}. During the first half of the gait cycle, the first group enters the swing phase, while the second group remains in contact with the ground, providing support. In the second half of the cycle, the groups exchange roles. 
In the presence of leg damage, an MLR has to adaptively transition from this nominal gait to an alternative stable gait pattern. This adaptation must ensure that at least three functional legs remain in the support phase at all times. Therefore, we propose to use modified versions of the quadrangular or pentagonal gaits for damages involving one or two leg loss, respectively. 
\begin{algorithm}[htbp]
\caption{Gait Sequence Planning Algorithm} \label{chap6:alg1}
\begin{algorithmic}[1]
\REQUIRE Morphology vector $\EX$
\STATE $\hat N= \sum\limits_{i=1}^{N} \ex_i$
\IF{$\hat N=6$}
    \STATE Use nominal gait
\ELSIF{$\hat N=5$}
    \STATE Run Algorithm \ref{chap6:alg3}
\ELSIF{$\hat N=4$ AND $\sum\limits_{i~ \text{even}}\ex_i>1$ AND $\sum\limits_{i~ \text{odd}}\ex_i>1$}
\STATE Run Algorithm \ref{chap6:alg4}
\ELSE
    \STATE Error!
\ENDIF
\RETURN New gait sequence 
\end{algorithmic}
\end{algorithm}

To make the gait reconfiguration process autonomous, we incorporated the ``rule of neighborhood" strategy \cite{ref16, porta2004reactive, song1987analytical} into our algorithm. According to this rule, if a leg enters the swing phase, both its neighboring legs must remain in the support phase. It has been analytically proven that any gait satisfying this rule is statically stable under some conditions, as investigated in \cite{suphi2007analysis}. Therefore, we employ the gait sequence planning in  Algorithm \ref{chap6:alg1}, given the morphology vector $\EX$ of a damaged MLR and the gait period $T_g$.  
Following this algorithm, the robot can autonomously switch from a nominal tripod gait to a modified quadrangular gait when one leg is damaged, and further adapt to a modified pentagonal gait when two legs are damaged. 

\begin{algorithm}[htbp]
\caption{Modified Quadrangular Sequence for Hexapod with One Damaged Leg} \label{chap6:alg3}
\begin{algorithmic}[1]
\REQUIRE Morphology vector $\EX$ and gait period $T_g$
\REQUIRE $\hat N=5$
    \STATE Set $\gamma=1$
    \FOR{$i \in\mathcal L_1$}
        \IF{$\ex_{i} = 1$}
            \STATE Set $\alpha_{\gamma}=i$
            \STATE Set $\gamma= \gamma+1$
        \ENDIF     
        \IF{$\gamma=3$}
            \STATE Set $\lambda = i$
        \ENDIF
    \ENDFOR
    \STATE Set $\gamma=1$
    \FOR{$i \in\mathcal L_2$}
        \IF{$\ex_{i} = 1$}
            \STATE Set $\beta_{\gamma}=i$
            \STATE Set $\gamma= \gamma+1$
        \ENDIF      
        \IF{$\gamma=3$}
            \STATE Set $\lambda = i$
        \ENDIF
    \ENDFOR
    \STATE Legs $\alpha_1,\beta_2$ in the \textbf{swing phase} from $t=0$ to $\frac{T_g}{3}$
    \STATE Other Legs in the \textbf{support phase} from $t=0$ to $\frac{T_g}{3}$
    \STATE Legs $\alpha_2,\beta_1$ in the \textbf{swing phase} from $t=\frac{T_g}{3}$ to $\frac{2T_g}{3}$
    \STATE Other Legs in the \textbf{support phase} from $t=\frac{T_g}{3}$ to $\frac{2T_g}{3}$
    \STATE Leg $\lambda$ in the \textbf{swing phase} from $t=\frac{2T_g}{3}$ to $T_g$
    \STATE Other Legs in the \textbf{support phase} from $t=\frac{2T_g}{3}$ to $T_g$
\RETURN Gait sequence 
\end{algorithmic}
\end{algorithm}
\begin{algorithm}[htbp]
\caption{Modified Pentagonal Sequence for Hexapod with Two Damaged Legs} \label{chap6:alg4}
\begin{algorithmic}[1]
\REQUIRE Morphology vector $\EX$ and gait period $T_g$
\REQUIRE $\hat N=4$
\STATE Set $k=0$
    \FOR{$i = 1$ to $ N$}
    \IF{$\ex_{i} = 1$}
        \STATE Leg $i$ in the \textbf{swing phase} from $t=k\frac{T_g}{4}$ to $(k+1)\frac{T_g}{4}$
        \STATE Other Legs in the \textbf{support phase} from $t=k\frac{T_g}{4}$ to $(k+1)\frac{T_g}{4}$
        \STATE Set $k = k + 1$
    \ENDIF
    \ENDFOR
\RETURN Gait sequence 
\end{algorithmic}
\end{algorithm}

\subsection{DE-based Leg Trajectory Planning}\label{chap6:DE}

In the second step of the proposed recovery strategy, we optimize the parameters shaping the leg tip trajectories within a gait. Each trajectory is designed relative to the robot’s main body to achieve straight-line locomotion along the $Y$-axis, as illustrated in Fig. \ref{chap6:Robot}, while maintaining the body parallel to the ground plane. Accordingly, the desired tip position of Leg $i$ in the main body frame, denoted by $\Bar{p}^b_{b,it}$, is defined by the following three position components:
\begin{align}   
    \begin{cases}
      \Bar{p}^b_{b,it_x} = x_{0_{i}} = constant\\
      ^{swing}\Bar{p}^b_{b,it_y} = y_{0_{i}} -0.5L_{sl}\cos\big(\frac{\sigma\pi}{N_g} (k - 1 ) \big) \\
       ~~~~~~~~~~~~~~~~~~~~~~~~~~~~~~~~~~~~~~~~~~~~~~~~~~k =  1,  \dotsb , \frac{N_g}{\sigma}\\
       ^{support}\Bar{p}^b_{b,it_y} = y_{0_{i}} +0.5L_{sl}\cos\big(\frac{\sigma\pi}{(\sigma -1)N_g} (k-\frac{N_g}{\sigma}-1) \big) \\
       ~~~~~~~~~~~~~~~~~~~~~~~~~~~~~~~~~~~~~~~~~~~k =  \frac{N_g}{\sigma} + 1,  \dotsb , N_g\\
       ^{swing}\Bar{p}^b_{b,it_z} = z_{0_{i}} + 0.5L_{sh}\bigg(1-\cos\big( (\frac{2\sigma\pi}{N_g})(k-1) \big) \bigg)- H_0 \\
       ~~~~~~~~~~~~~~~~~~~~~~~~~~~~~~~~~~~~~~~~~~~~~~~~~~k = 1,  \dotsb  ,\frac{N_g}{\sigma}  \\
      ^{support}\Bar{p}^b_{b,it_z} = z_{0_{i}} + 0.5L_{sd} \\ ~~~~~~~~~~~~~~~~~\bigg(1-\cos\big( (\frac{2\sigma\pi}{(\sigma - 1)N_g})(k-\frac{N_g}{\sigma}-1) \big) \bigg) - H_0 \\
      ~~~~~~~~~~~~~~~~~~~~~~~~~~~~~~~~~~~~~~~~~~~k = \frac{N_g}{\sigma} + 1,  \dotsb , N_g
    \end{cases}
    \label{chap6:tip_ref}
\end{align}
Here, we have $\sigma=2,3,4$ for the tripod, modified quadrangular, and modified pentagonal gait sequences, respectively.
The parameters $L_{sl}$, $L_{sh}$, and $L_{sd}$ denote the step length, height, and depth, respectively, and $H_0$ is the main body's initial height. The components of the initial position of the tip of Leg $i$ in the main body frame is denoted by $x_{0_{i}}$, $y_{0_{i}}$, and $z_{0_{i}}$. Based on a time step $\Delta t$ and the gait period $T_g$, we  defined $N_g=T_g/\Delta t$ to be the number of points on the gait trajectory. The leg tip trajectory is used to derive the joint trajectories using standard numerical or analytical inverse kinematics methods, as each leg is considered a fixed-base manipulator during the trajectory design process. The controller aims to control each leg relative to the body in joint space.


In this letter, we develop a DE algorithm to recover the ability of a damaged MLR to locomote. DE is a population-based stochastic optimization algorithm introduced by Storn and Price \cite{storn1997differential}. It is widely used for solving complex optimization problems due to its simplicity, robustness, and efficiency.
Before detailing the proposed optimization procedure, we define the optimization variables and objective function. According to \eqref{chap6:tip_ref}, we can meaningfully alter the shape of a tip trajectory by modifying two parameters, $x_{0_{i}}$ and $y_{0_{i}}$, for each leg, and two parameters used for all legs, $L_{sl}$ and $L_{sh}$, totaling $D=14$ optimization variables for a hexapod: 
\begin{align}
    \X = [y_{0_{1}} ~~ ...~~y_{0_{6}}~~x_{0_{1}}~~...~~x_{0_{6}} ~~L_{sl}~~L_{sh}] \in \mathbb{R}^{D}.
\end{align}
Variation of the other parameters, i.e., $z_{0_i}$, $L_{sd}$, and $H_0$, were tested, showing negligible impact on walking performance. We impose realistic bounds to define the feasible set of optimization variables. For $x_{0_i}$,  these bounds are determined by the robot’s body width and height and leg link lengths. To prevent leg-to-leg collisions during gait cycle, dynamic constraints based on $y_{0_i}$ and $L_{sl}$ are incorporated. In the event of a leg damage, these constraints are selectively relaxed for neighboring legs, enabling them to either extend their step length or provide a contact point close to that of the damaged leg. Accordingly, we denote the set of all feasible candidate recovery solutions by $\mathcal{K}$. 
A key challenge is formulating an appropriate objective function that effectively evaluates the locomotion capability of the MLR. Building on multiple objective functions examined in \cite{kon2020gait}, we develop the following objective function to be maximized:
\begin{align}\label{obj-func}
         \mathcal{F}(\X)= \Big( w_1 y_{f}^2 \Big)/ \Big( 1 + w_2 x_{f}^2 + w_3 \psi_{f}^2
         + 
          w_4 \Delta \phi^2 + w_5 \Delta \theta^2 \Big), 
\end{align}
where we simulate a damaged MLR with the candidate $\X$ for a fixed number of gait cycles to evaluate $x_f$ and $y_f$ being the final position of the MLR's CoM in the $X$ and $Y$ directions. We use ZYX Euler angles to parameterize the orientation of the main body using roll ($\phi$), pitch ($\theta$), and yaw ($\psi$) angles. Accordingly, $\psi_f$ is the final yaw angle and $\Delta\phi$ and $\Delta\theta$ are respectively the maximum oscillation amplitude of the roll and pitch angles of the main body during the simulation. Finally $w_1,\ldots,w_5$ are the weights indicating the contribution of each term in $\mathcal{F}$.
We claim that a large $\mathcal{F}$ corresponds to a candidate tip trajectory that can recover the MLR's locomotion capability. Therefore, the optimizatoin problem we solve in this step of gait reconfiguration is formulated as: 
\begin{align}
         \X_{best}&=\arg\!\max_{\X \in \mathcal{K}} \mathcal{F}. \label{DE:fitness} 
\end{align}
We use a DE algorithm with population size $P_s$ and generation size $G_s$ to iteratively solve this problem using mutation, crossover, and selection operations, as detailed step-by-step in the following:

\subsubsection{Initialization}
Randomly generate an initial population of unique candidates $\{{}^1\X^1,\ldots,{}^1\X^{P_s}\}$  uniformly distributed in the feasible search space $\mathcal{K}$.

\subsubsection{Evaluation}

Once a population is generated, each candidate is fed into the simulation engine \cite{farghdani2024singularity,future:farghdani_model2} to produce the trajectory of the main body's orientation and CoM position over a period of time. Then, we evaluate quantities $x_f,y_f,\psi_f,\Delta\phi,$ and $\Delta\theta$ to calculate the objective function in \eqref{obj-func} for each candidate.

\subsubsection{Mutation}
For the candidate $k$ of the population in the generation $g$, ${}^g\X^k$, a mutant vector $^{g+1}\V^k \in\R^D$ is generated using the current-to-best strategy to balance exploration and exploitation:
    \begin{equation}
        ^{g+1}\V^k = \ ^g\X^k + K_f  (^g\X^{best} -\ ^g\X^{k}) + K_f  (\ ^g\X^{r1} - \ ^g\X^{r2})
    \end{equation}
    where $^g\X^{best}$ is the best performing solution in the current generation, $^g\X^{r1},\ ^g\X^{r2}$ are randomly selected distinct candidates from the population in generation $g$, and $K_f \in [0,1]$ is the mutation constant.

\subsubsection{Crossover}
A trial vector $^{g+1}\U^k \in\R^D$ is created by combining the mutant vector $^{g+1}\V^k$ and the candidate $^g\X^k$ using binomial crossover:
    \begin{equation}
        ^{g+1}\U^{k}_m = \begin{cases} 
            ^{g+1}\V^{k}_m, & \text{if } r_c^m \leq CR \\
            ^g\X^{k}_m, & \text{otherwise}
        \end{cases}
    \end{equation}
where the subscript $m\in\{1,\ldots,D\}$ denotes the element of a vector, $r_c^m$ is a randomly generated number between zero and one, and $CR \in [0,1]$ is the crossover probability.

\subsubsection{Constraint violation}
Before evaluating the objective function at the new candidates, the violation of the constraints should be checked. Therefore, we refine the newly generated candidates according to the following rule:
    \begin{equation}
        ^{g+1}\U^{k}_m = \begin{cases} 
            \X^{max}_{m}, & \text{if } \ ^{g+1}\U^{k}_m > \X^{max}_{m} \\
            \X^{min}_{m}, & \text{if } \ ^{g+1}\U^{k}_m < \X^{min}_{m} \\
            ^{g+1}\U^{k}_m, & \text{otherwise}
        \end{cases}
    \end{equation}
    where for $m=1,\ldots,D$, the bounds $\X^{max}_m$, and $\X^{min}_m$ are the maximum and minimum acceptable values for each variable in the candidate solution, respectively.

\subsubsection{Selection}The next generation is now formed by applying the selection rule:
    \begin{equation}
        ^{g+1}\X^k = \begin{cases} 
            ^{g+1}\U^k, & \text{if } \mathcal{F}(\, ^{g}\X^k)  \leq \mathcal{F}(\, ^{g+1}\U^k) \\
            ^{g}\X^k, & \text{otherwise}
        \end{cases}
    \end{equation}
    where $\mathcal{F}$ is the objective function to be maximized. 

\subsubsection{Termination}
The DE optimization process repeats until a stopping criterion is met. This could be the maximum number of iterations $G_s$ or a convergence criterion.

\begin{figure}[hbt!]
  \centering
  \includegraphics[width=.35\textwidth]{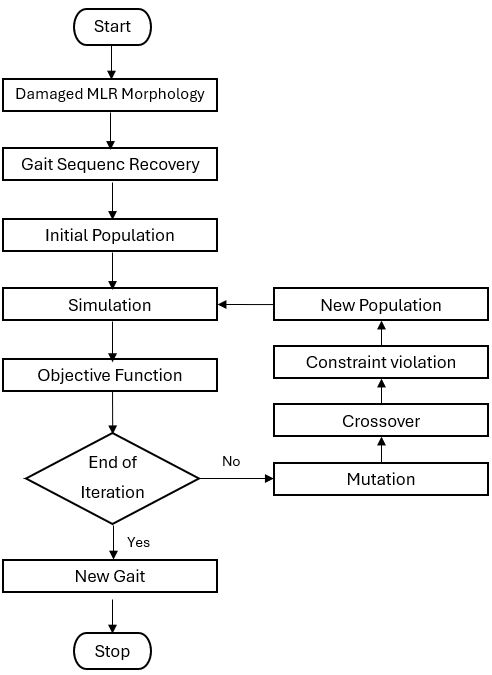}\\
  \caption{Damage recovery algorithm flowchart}
  \label{fig:DE}
\end{figure}

An overview of the damage recovery algorithm incorporating the gait sequence stabilizer and DE leg tip trajectory optimization is illustrated in Fig.~\ref{fig:DE}.

\section{Results \& Discussion}\label{chap6:results}

We evaluate the performance of the proposed damage recovery method for MLRs using the gait reconfiguration algorithm described in Section~\ref{chap6:method}. The effectiveness of the approach is demonstrated through a series of damage scenario experiments, covering both single- and double-leg failures, using a real six-legged robot. These experiments incorporate both internal sensing and external motion tracking, as detailed in the following sections.

\subsection{System Description}

The experimental setup includes a Hiwonder JetHexa, a six-legged robot with 3-Dof legs (see Fig. \ref{chap6:Robot}). Each leg has three serial bus servo motor-actuated joints. The onboard robot computer is an NVIDIA Jetson Nano Developer Kit with an MPU6050 inertial measurement unit (IMU) \cite{jethexa}. This IMU is used to obtain the roll, pitch, and yaw of the robot's body, and motion-tracker system is used to obtain both the main body orientation and translation during the experiments.
In all our experimental tests and simulations, we utilized the model and control structure introduced in \cite{farghdani2024singularity, future:farghdani_model2}. 

\subsection{DE Algorithm Setup}

To execute the DE algorithm, the evaluation step involves computing the desired joint trajectories corresponding to the leg tip trajectories generated from \eqref{chap6:tip_ref}. These joint trajectories are obtained using the closed-form inverse kinematics solution for a 3-DoF leg, as described in \cite{8324813}.
\begin{align}    
    \begin{cases}
      \Bar{q}_{i1} =&\arctan(\Bar{p}^b_{b,it_y}/\Bar{p}^b_{b,it_x})\\
      \Bar{q}_{i2} =&\arcsin(\Bar{p}^b_{b,it_z}/\sqrt{(L_{i3} c_{i3} + L_{i2})^2 + (L_{i3} s_{i3})^2}) \\
      &- \gamma_i\\
      \Bar{q}_{i3} =&\arcsin( ((c_{i1} \Bar{p}^b_{b,it_x} + s_{i1} \Bar{p}^b_{b,it_y} - L_{i1})^2 \\
      &+ (\Bar{p}^b_{b,it_z})^2 - L_{i3}^2 - L_{i2}^2 )/ (2 L_{i2} L_{i3}) ) - \pi/2
    \end{cases}
\end{align}
where $\gamma_i  \equiv \arctan ( L_{i3} s_{i3} / (L_{i3} c_{i3} + L_{i2}) )$; $c_{ij}  \equiv \cos(q_{ij})$; $s_{ij}  \equiv \sin(q_{ij})$, and $L_{ij}$ is the length of the link $j$ of Leg $i$. A joint space PID controller is then designed to follow these trajectories with $K^p_{ij} = 500$, $K^d_{ij} = 1$, and $K^I_{ij} = 10$ gains for all joints. 
In addition, several parameters need be specified for the DE algorithm whose values are summarized in Table~\ref{table:DE}.

\begin{table}[ht]
\caption{Damage identification algorithm parameters} 
\color{black}
\centering 
\begin{tabular}{c c c c c } 
\hline\hline 
  Parameter & Value &  Parameter & Value\\ 
\hline 
$P_s$ & 30 & $G_s$ & 60 \\
$CR$ & 0.6 & $K_f$ & 0.5 \\
Simulation time & 10 sec &  $w_1$ & 1 \\
$w_2$ & 1 & $w_3$ & 10 \\
$w_4$ & 100 & $w_5$ & 100 \\
\\
\hline 
\end{tabular}
\label{table:DE} 
\end{table}
To define the feasible set of optimization variables, we introduce the maximum and minimum acceptable bounds for solutions:
\begin{align*}\nonumber
    \X^{max} = &(0.05~~0.05~~ 0.03~~ 0.03~~ 0.01~~ 0.01~~ 0.13 \\
                &~~~~0.13~~ 0.13 ~~0.13~~ 0.13~~ 0.13~~ 0.05~~ 0.05) \\ \nonumber
    \X^{min} = &(-0.01~-0.01~-0.03~-0.03~-0.05~-0.05\\
                &~~~~0.09~~0.09 ~~0.09~~ 0.09~~ 0.09~~ 0.09~~ 0.02~~ 0.02)
\end{align*}
As explained in Section~\ref{chap6:DE}, the bounds on $y_{0_{i}}$ are relaxed by $20\%$ for the legs adjacent to a damaged leg, allowing for increased adaptability in response to the effects of damage.

\subsection{Experimental Results}
\begin{figure}[hbt!]
  \centering
  \includegraphics[width=.35\textwidth]{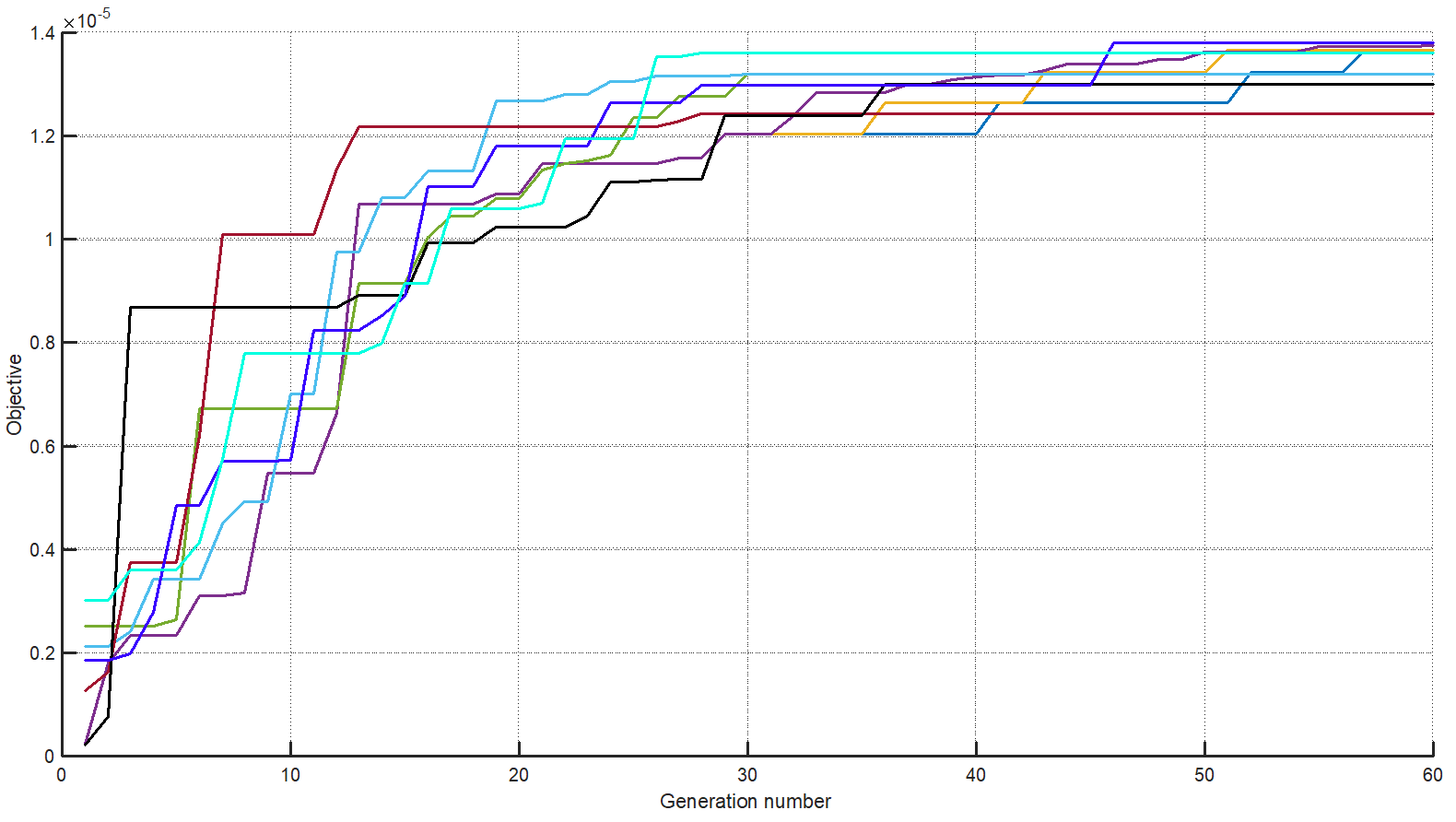}\\
  \caption{The best objective function value per generation for the scenario where Legs 1 and 6 are missing}
  \label{chap6:j_legs1_6}
\end{figure}

\begin{figure}[hbt!]
  \centering
  \includegraphics[width=.45\textwidth]{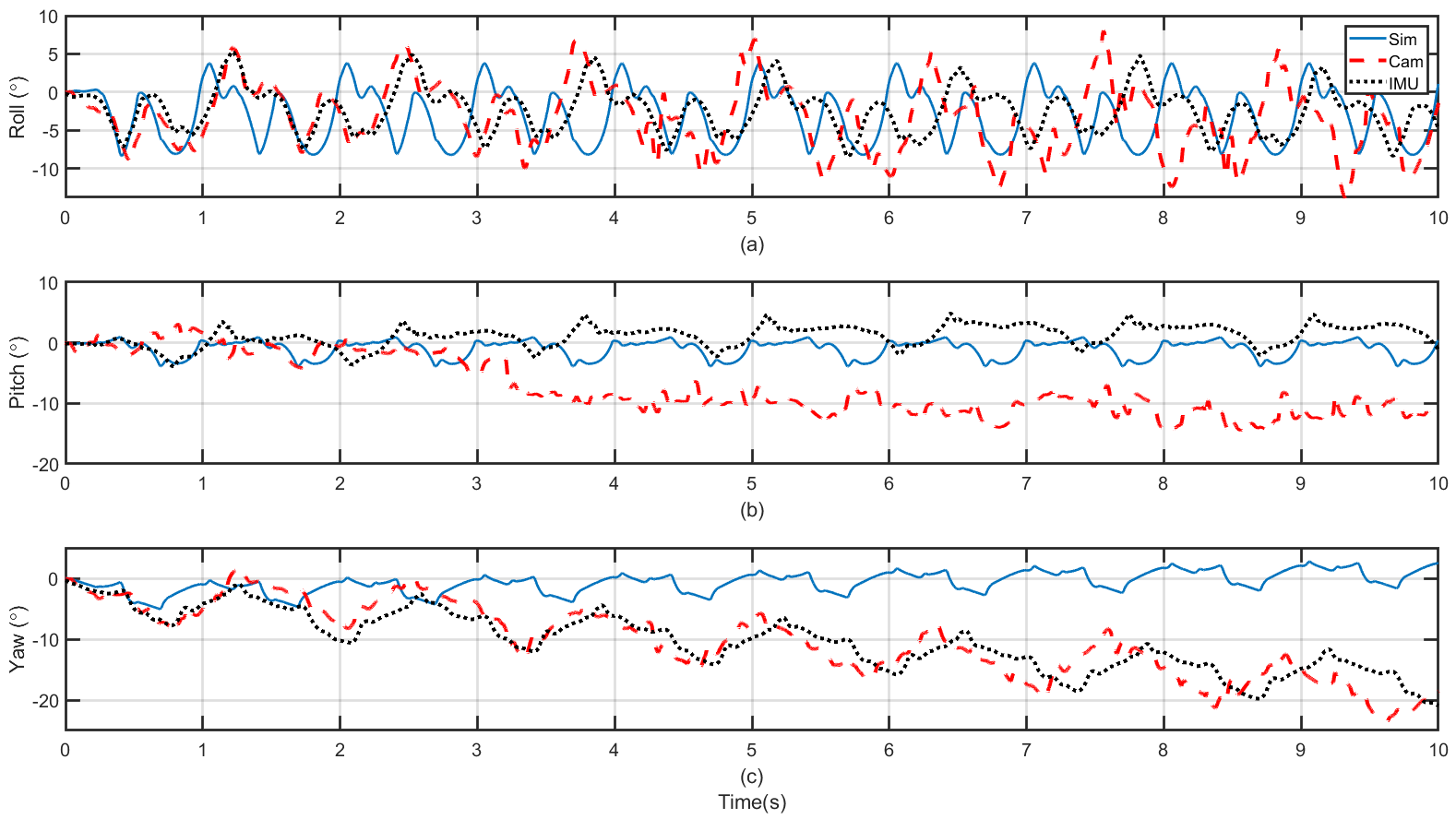}\\
  \caption{Main body orientation during the scenario where Legs 1 and 6 are missing}
  \label{chap6:legs1_6_o}
\end{figure}

\begin{figure}[hbt!]
  \centering
  \includegraphics[width=.45\textwidth]{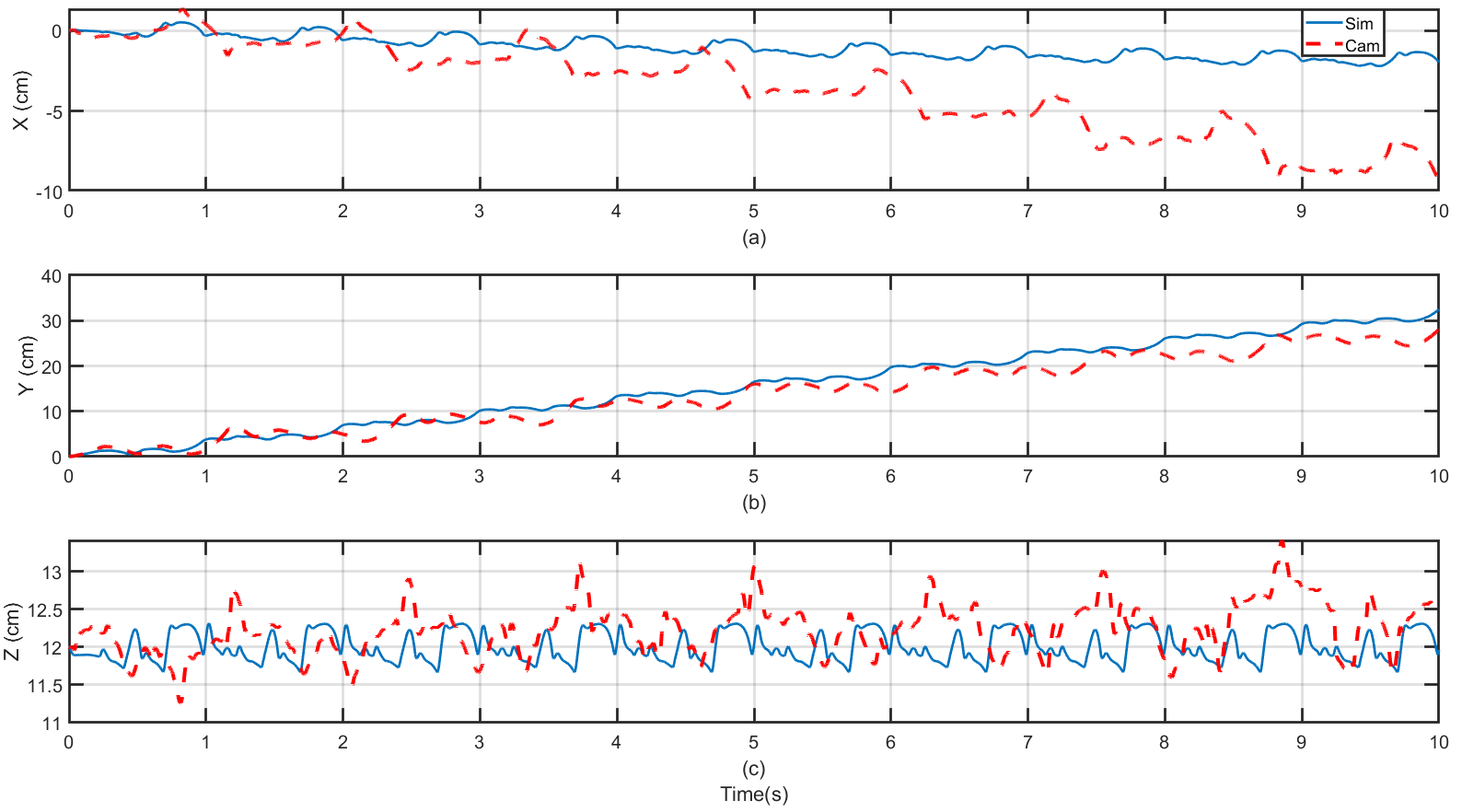}\\
  \caption{Main body COM position during the scenario where Legs 1 and 6 are missing}
  \label{chap6:legs1_6_p}
\end{figure}

\begin{figure}[hbt!]
  \centering
  \includegraphics[width=.35\textwidth]{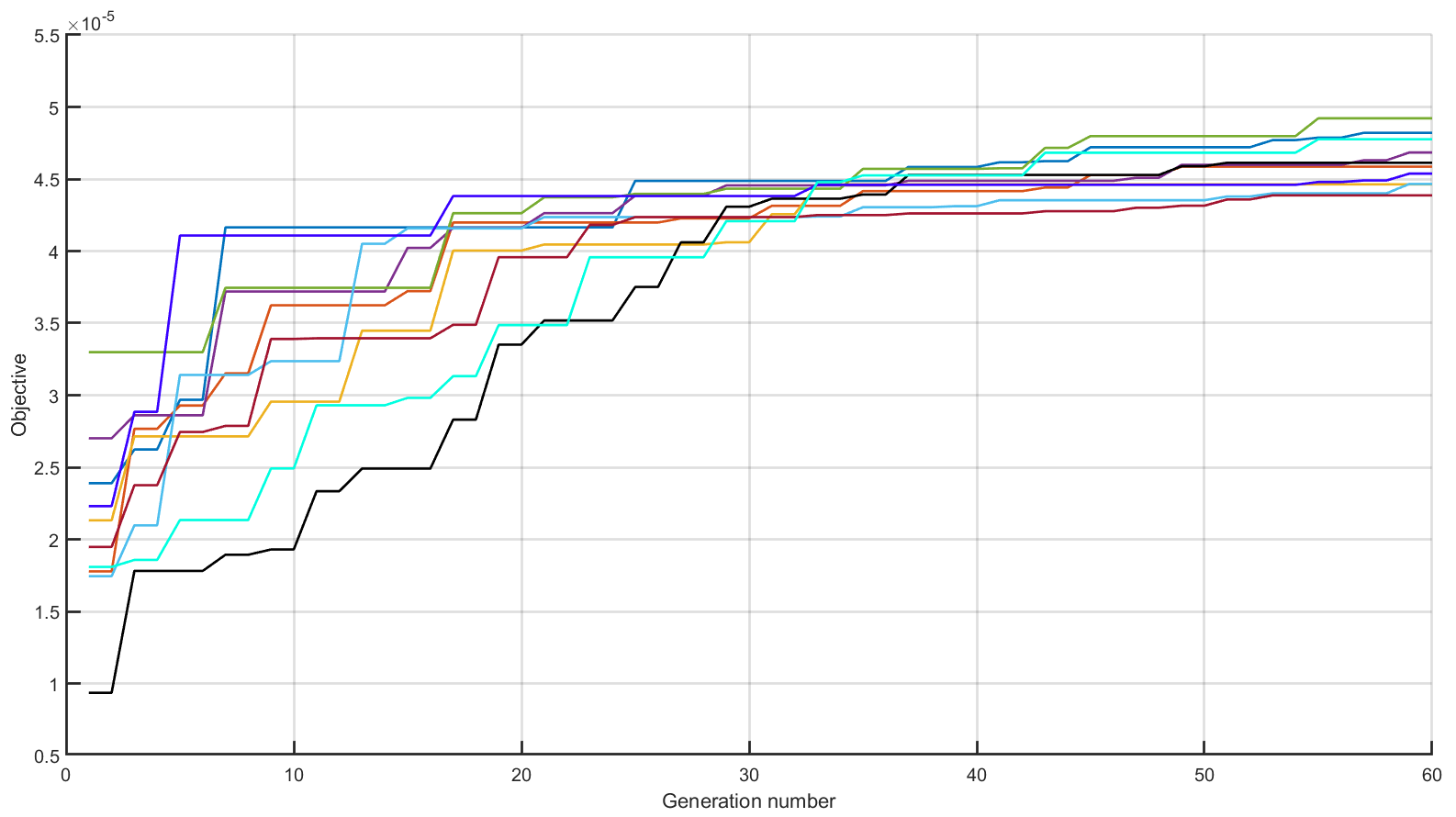}\\
  \caption{The best objective function value per generation for the scenario where Legs 3 and 4 are missing}
  \label{chap6:j_legs3_4}
\end{figure}

\begin{figure}[hbt!]
  \centering
  \includegraphics[width=.45\textwidth]{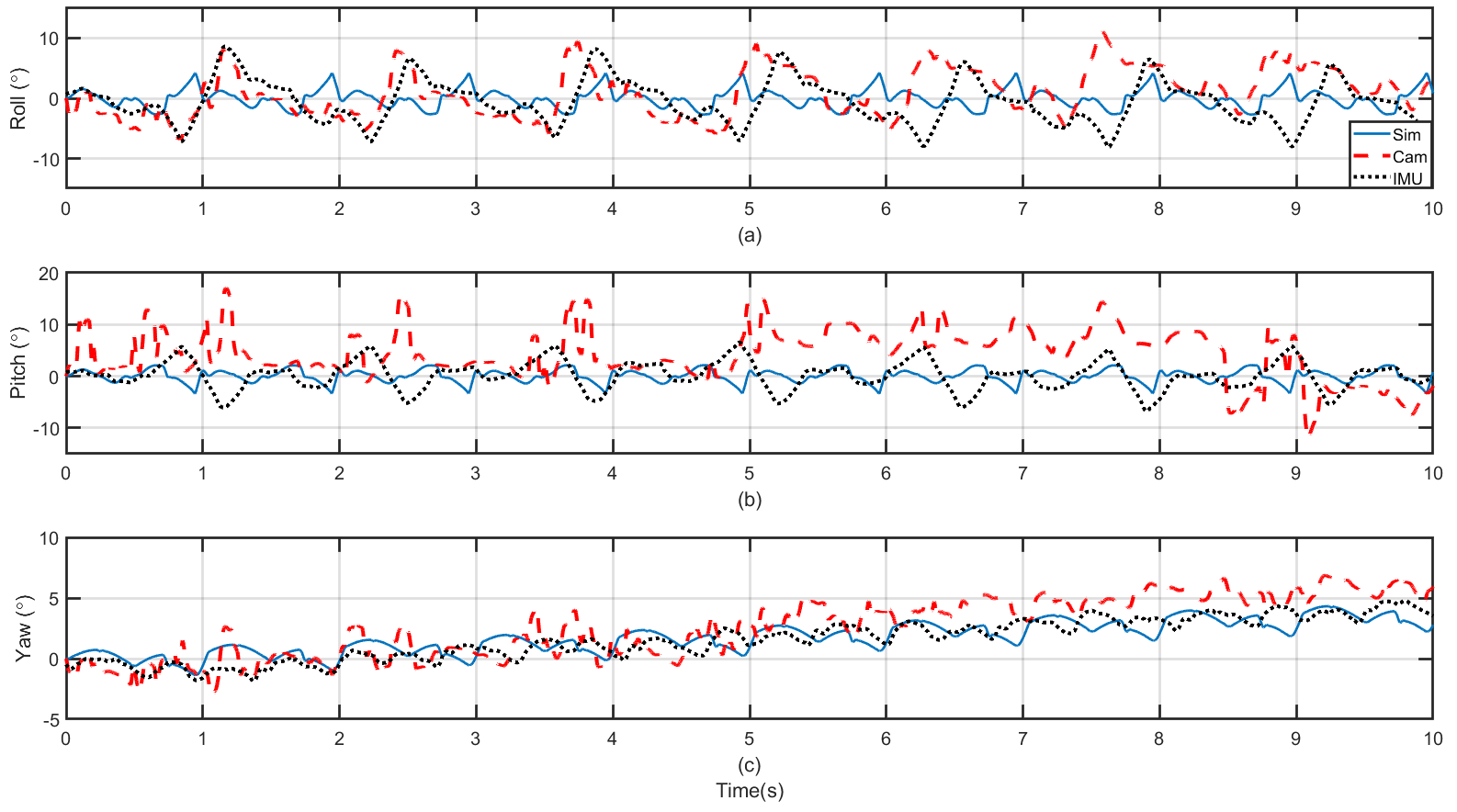}\\
  \caption{Main body orientation during the scenario where Legs 3 and 4 are missing}
  \label{chap6:legs3_4_o}
\end{figure}

\begin{figure}[hbt!]
  \centering
  \includegraphics[width=.45\textwidth]{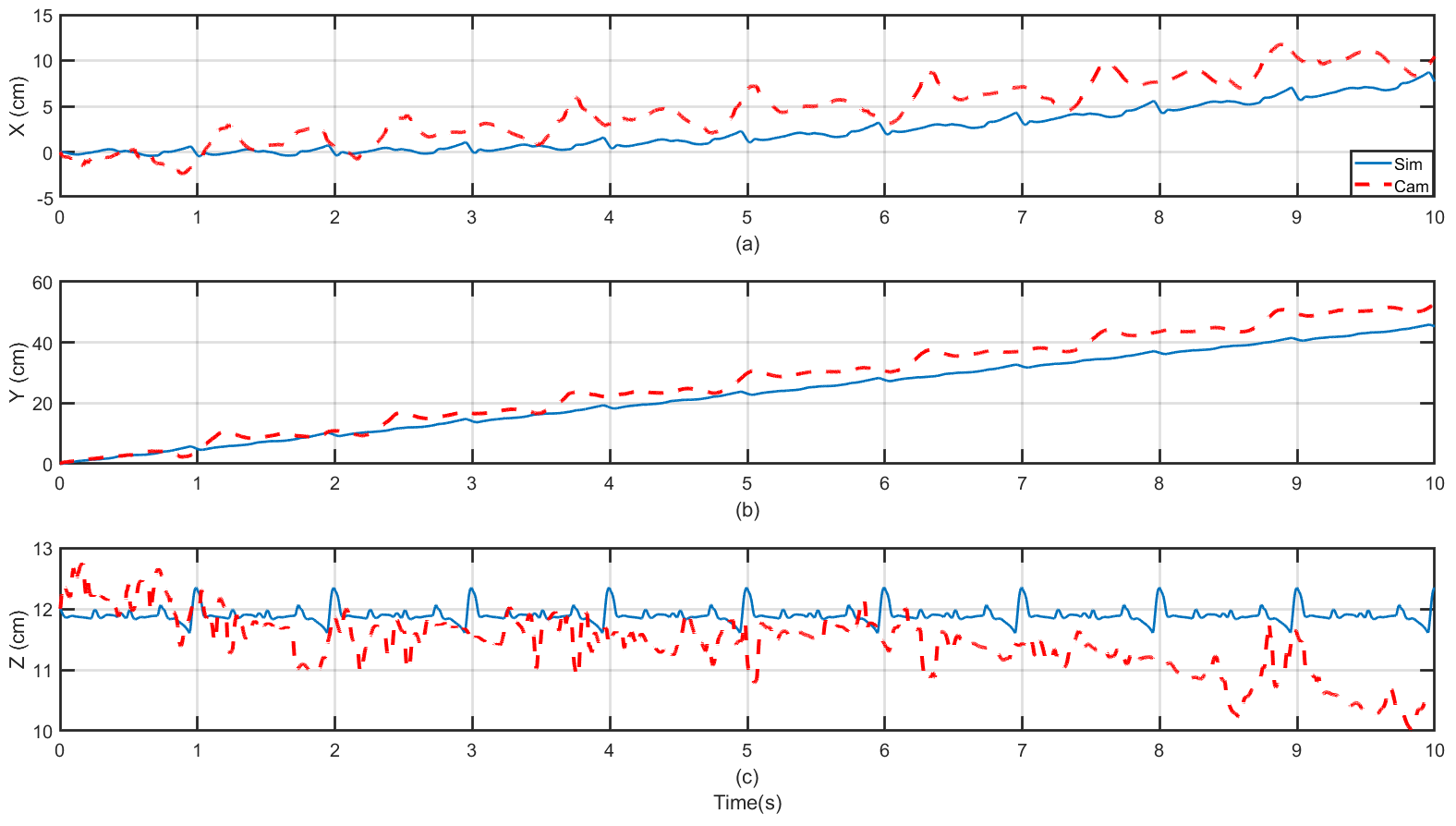}\\
  \caption{Main body COM position during the scenario where Legs 3 and 4 are missing}
  \label{chap6:legs3_4_p}
\end{figure}

\begin{figure}[hbt!]
  \centering
  \includegraphics[width=.35\textwidth]{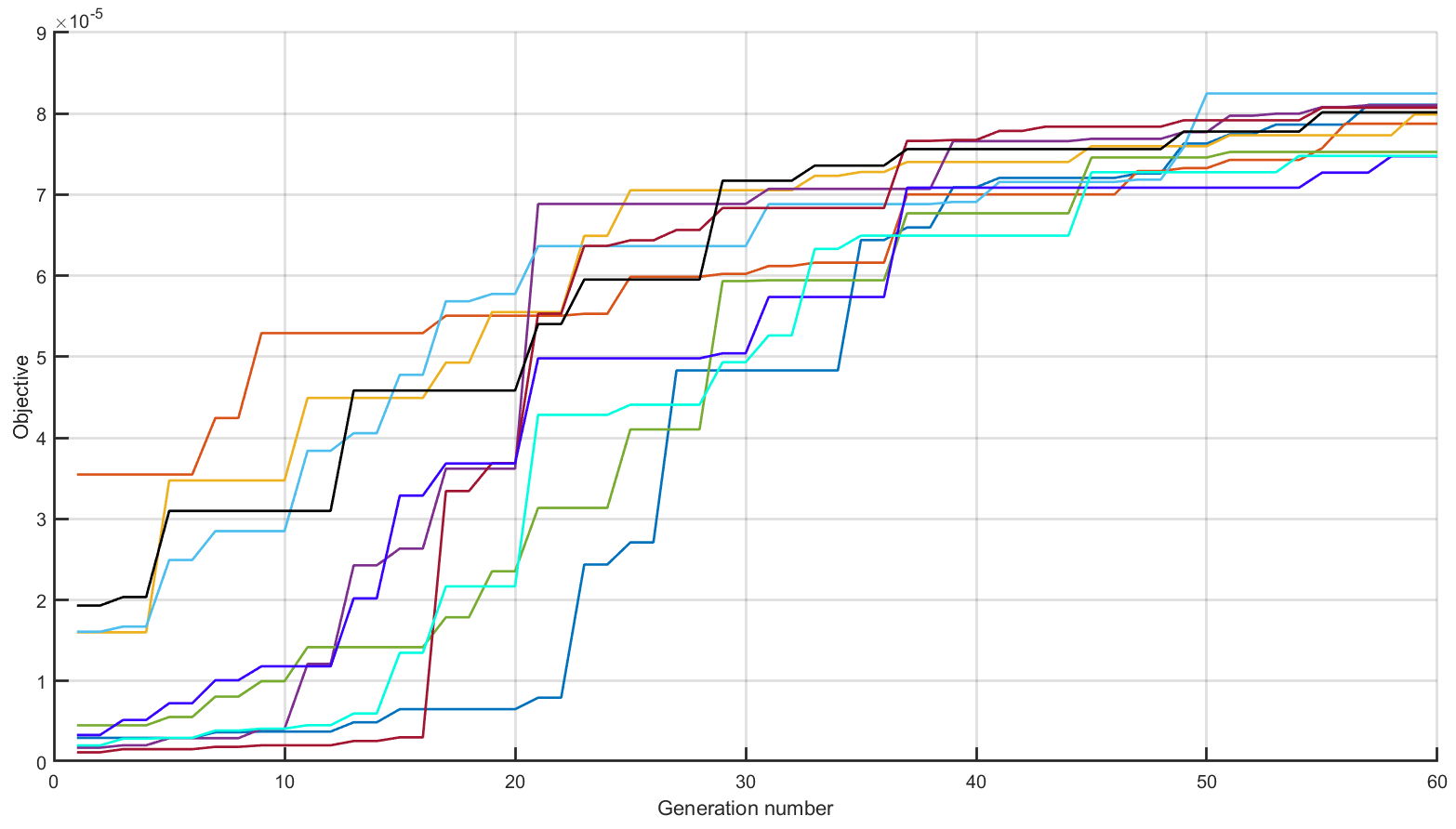}\\
  \caption{The best objective function value per generation for the scenario where Leg 1 is missing}
  \label{chap6:j_leg1}
\end{figure}

\begin{figure}[hbt!]
  \centering
  \includegraphics[width=.45\textwidth]{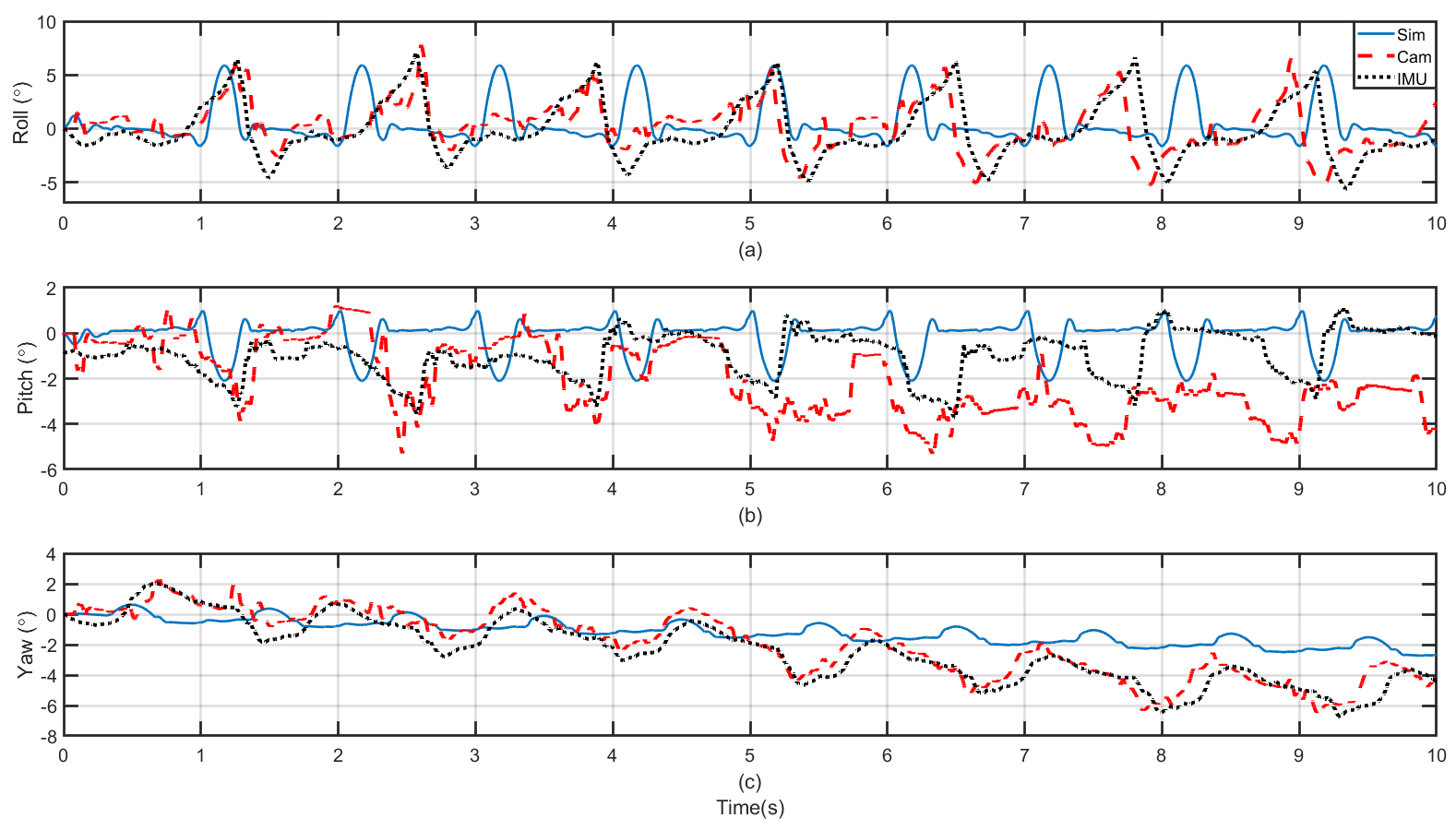}\\
  \caption{Main body orientation during the scenario where Leg 1 is missing}
  \label{chap6:leg1_o}
\end{figure}

\begin{figure}[hbt!]
  \centering
  \includegraphics[width=.45\textwidth]{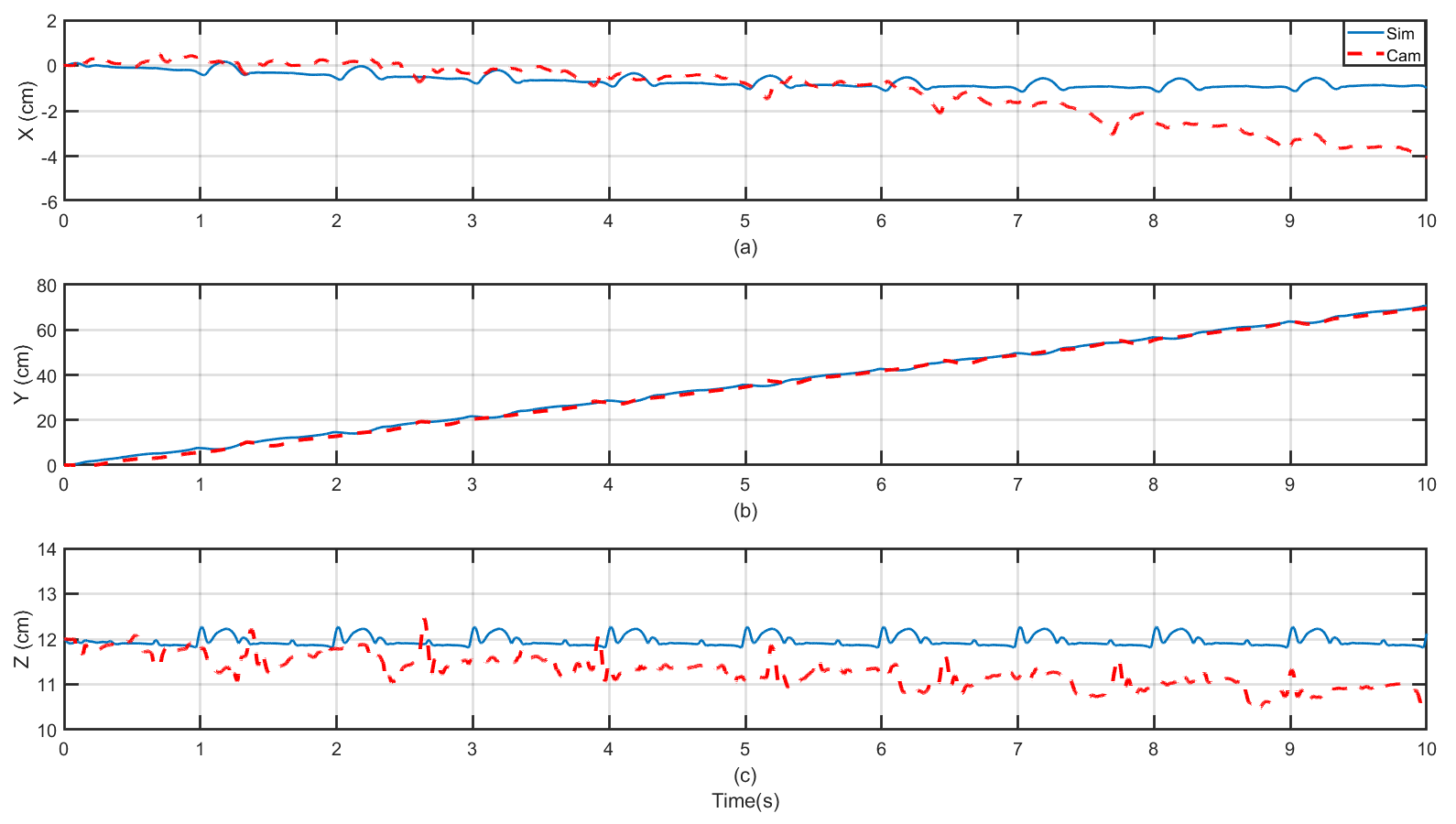}\\
  \caption{Main body COM position during the scenario where Leg 1 is missing}
  \label{chap6:leg1_p}
\end{figure}

\begin{figure}[hbt!]
  \centering
  \includegraphics[width=.35\textwidth]{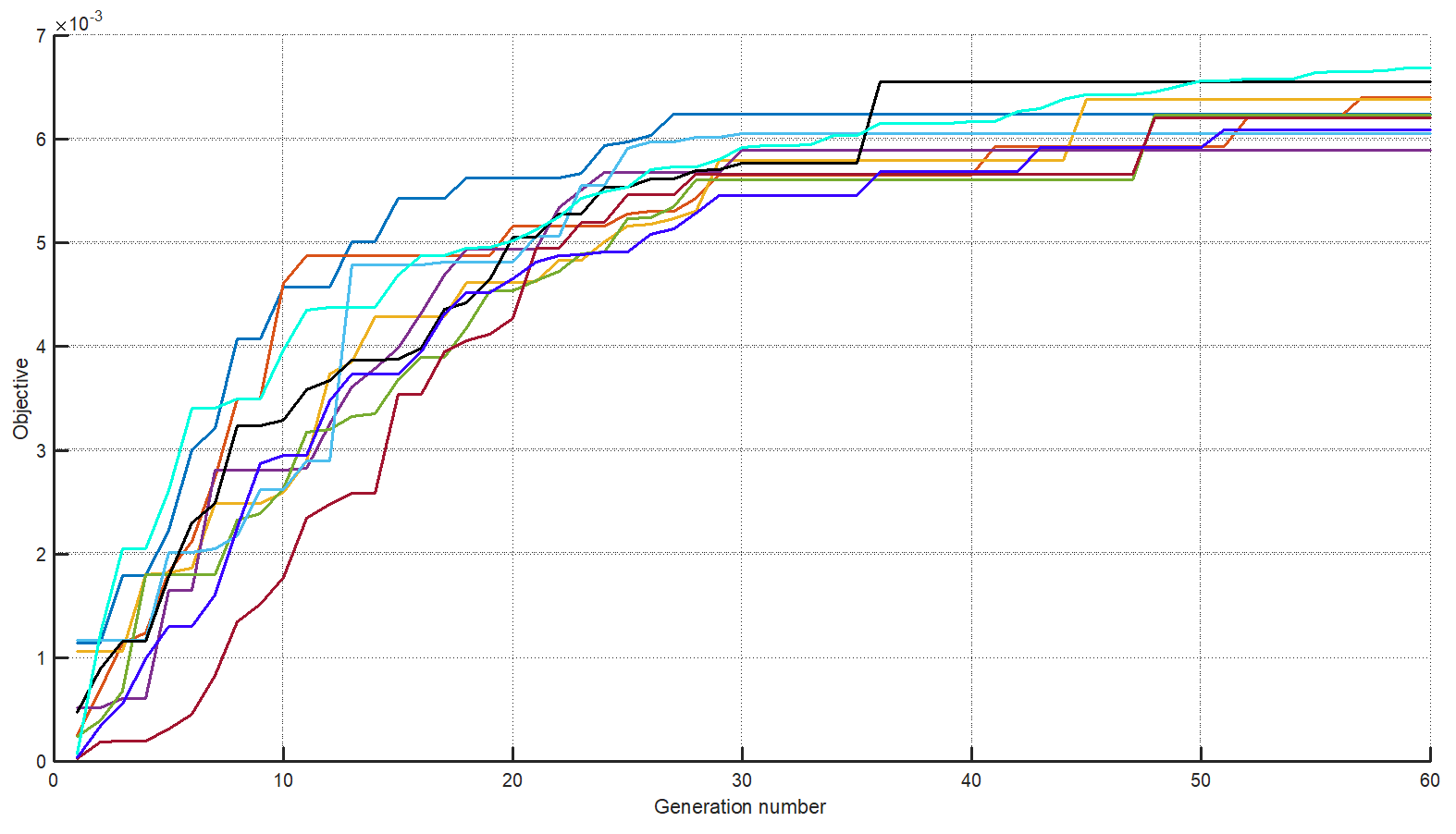}\\
  \caption{The best objective function value per generation for the scenario where Leg 3 is missing}
  \label{chap6:j_leg4}
\end{figure}

\begin{figure}[hbt!]
  \centering
  \includegraphics[width=.45\textwidth]{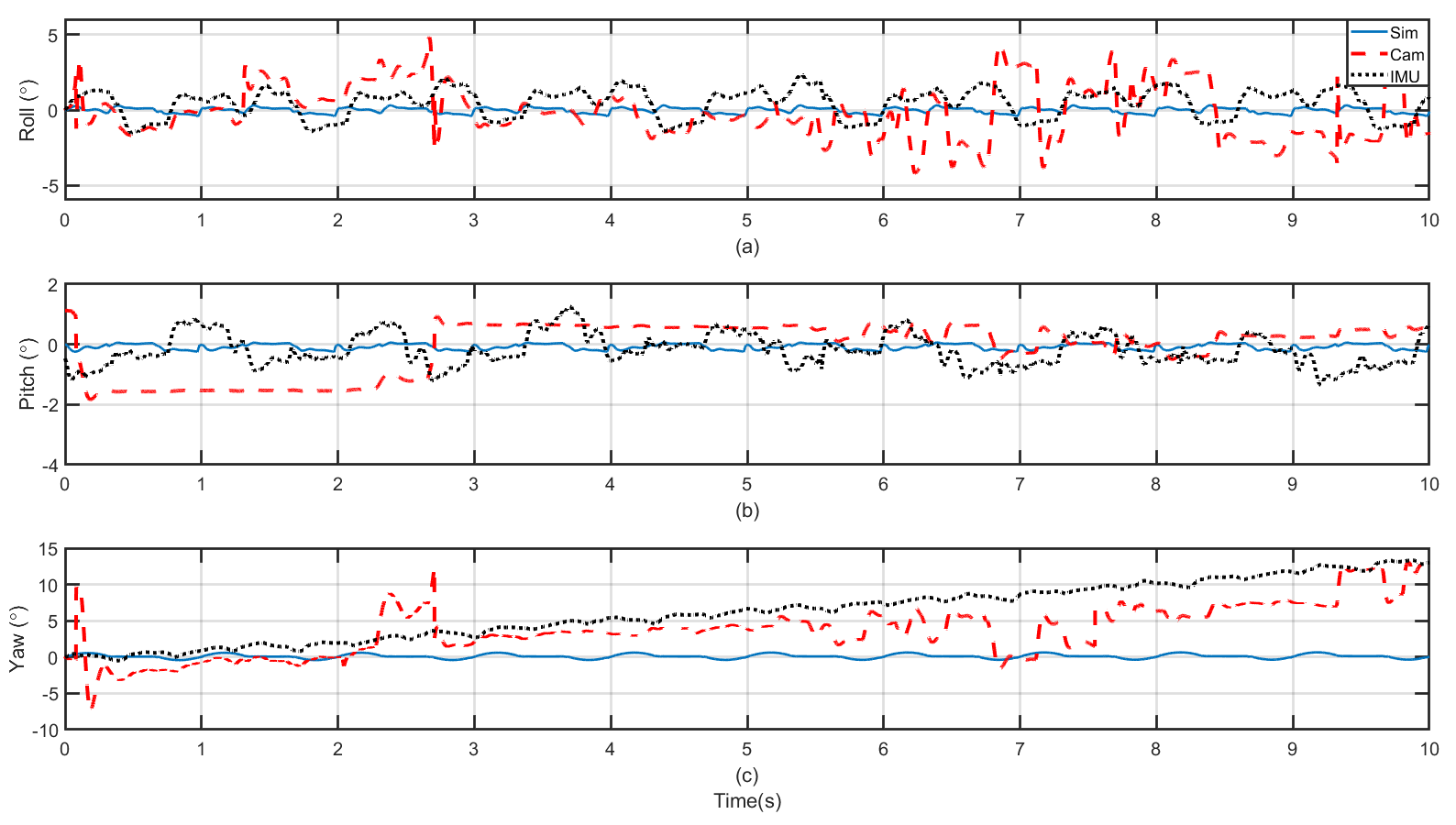}\\
  \caption{Main body orientation during the scenario where Leg 3 is missing}
  \label{chap6:leg4_o}
\end{figure}

\begin{figure}[hbt!]
  \centering
  \includegraphics[width=.45\textwidth]{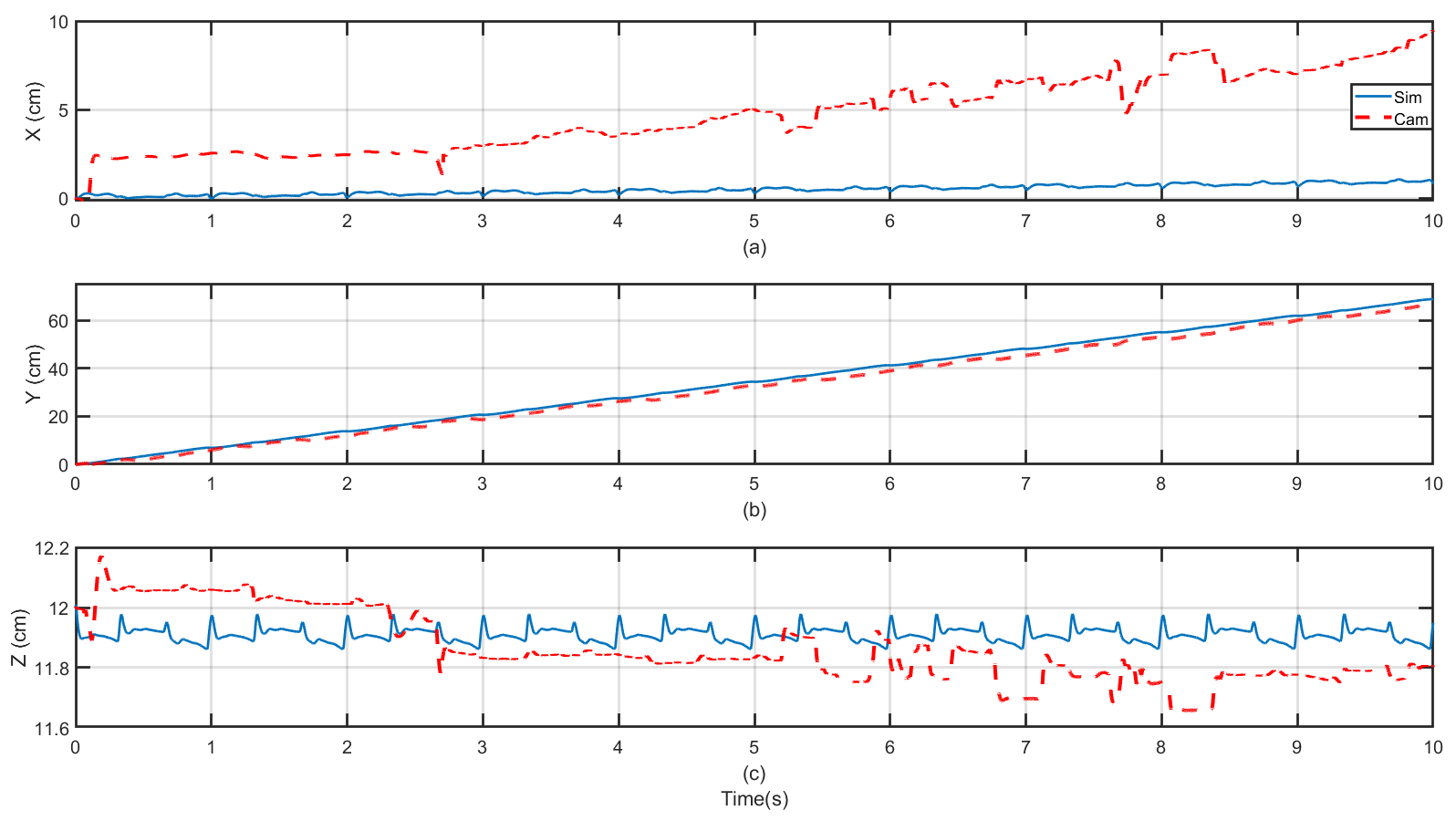}\\
  \caption{Main body COM position during the scenario where Leg 3 is missing}
  \label{chap6:leg4_p}
\end{figure}

In this section, we use the gait reconfiguration strategy detailed in this letter to recover the straight motion in a damaged hexapod under four different damage scenarios that are listed in Table \ref{chap6:tab_results}. 
\begin{table}[h!]
\centering
\caption{Summary of Damage Scenarios}
\label{chap6:tab_results}
\resizebox{\columnwidth}{!}{
\begin{tabular}{|l|ccc|ccc|}
\toprule
\textbf{Damage} &  & \textbf{Before Recovery} &  &  &\textbf{After recovery}&  \\ 
\textbf{Scenario} & \textbf{x (cm)} & \textbf{y (cm)} & \textbf{yaw ($^{\circ}$)} & \textbf{x (cm)} & \textbf{y (cm)} & \textbf{yaw ($^{\circ}$)} \\ 
\midrule
Legs 1 \& 6 missed    & N/A  & N/A  & N/A & -9.3 & 28.1 & -19.5  \\ 
Legs 3 \& 4 missed    & -2.3  & 1.5  & 2.4 & 10.1 & 52.7 & 5.8  \\ 
Leg 1 missed    & N/A  & N/A  & N/A & -4.1 & 69.5 & -4.3  \\ 
Leg 4 missed    & 14.6  & 26.6  & 64.0 & 9.4 & 66.9 & 13.0  \\ 
\bottomrule
\end{tabular}
}
\end{table}
For each damage scenario, we run our gait reconfiguration strategy 10 times to demonstrate robustness and convergence of the algorithm. Figures \ref{chap6:j_legs1_6}, \ref{chap6:j_legs3_4}, \ref{chap6:j_leg1}, and \ref{chap6:j_leg4} show the best objective function value per generation for all damage scenarios, demonstrating the convergence in all cases after almost 30 generations. To assess the algorithm's effectiveness, each best solution corresponding to a damage scenario is implemented on the damaged robot in the lab environment. We compare the predicted simulation result with real-world data obtained from the robot's IMU and the motion-tracker system. Robot orientation was observed using both IMU and motion-tracker data, while translational motion was evaluated exclusively using motion-tracker recordings due to the high noise typically associated with IMU-based linear acceleration measurements. The summary of the straight motion recovery performance (averaged over 10 best solutions) is depicted in Table \ref{chap6:tab_results}. In addition to that, Figures \ref{chap6:legs1_6_o} and \ref{chap6:legs1_6_p} (damage: Legs 1 \& 6 missing), \ref{chap6:legs3_4_o} and \ref{chap6:legs3_4_p} (damage: Legs 3 \& 4 missing), \ref{chap6:leg1_o} and \ref{chap6:leg1_p} (damage: Leg 1 missing), \ref{chap6:leg4_o} and \ref{chap6:leg4_p} (damage: Leg 4 missing), illustrate the motion of the main body's orientation and CoM position when a best solution is implemented on the damaged robot. 

To analyze the effectiveness of the damage recovery algorithm, we evaluated both the translational and rotational motions of the robot after gait reconfiguration. In terms of translation, in all scenarios the robot primarily moved in the desired forward direction (y-axis) with minimal lateral drift along the x-axis, aligning with our recovery objectives. The height variation (z-axis) remained small, indicating stable body motion throughout the tests. However, the robot's translational performance was slightly lower than expected compared to simulation results. This discrepancy can be attributed to modeling limitations, such as a simplified friction model, unmodeled system dynamics, and control challenges such as joint controller saturation.

We used ZYX Euler angles to track the robot’s orientation. The roll (y-axis) and pitch (x-axis) angles exhibited small oscillations with amplitudes below 20 degrees, which is within an acceptable range. The yaw angle (z-axis) was kept minimal, with total rotation under 20 degrees during the 10-second motion period, meeting the algorithm’s design goals. Nevertheless, yaw deviation was higher in real-world tests compared to simulations, likely due to factors such as CoM imbalance, leg backlash, and the aforementioned modeling inaccuracies.

For all tested damage scenarios, the objective function consistently converged within 30 generations, highlighting the algorithm’s efficiency and reliability in finding viable recovery strategies. Despite the performance gap between simulation and experimental tests, the real robot's behavior remained within acceptable limits, demonstrating the practical effectiveness of the proposed recovery algorithm. A summary of the robot’s motion in the x and y directions, along with the final yaw angle before and after recovery, is presented in Table \ref{chap6:tab_results}. Notably, in the scenarios where Legs 1 \& 6 were missing or Leg 1 alone was missing, the robot became unstable and fell before recovery, making pre-recovery results unavailable for comparison. This demonstrates the algorithm’s capability to restore locomotion even in extreme failure scenarios where the robot would otherwise be immobilized.



\section{Conclusion}\label{chap6:conclusion}

This letter presented a damage recovery algorithm for multi-legged robots, leveraging a fast and modular symbolic dynamic model. The proposed method restores locomotion in under an hour without requiring pre-training and demonstrates high success rates across diverse damage scenarios. A key aspect of the approach is the decomposition of recovery into stabilization and gait optimization, which effectively reduces path deviation and enhances stability. The algorithm was validated on a hexapod robot using both onboard IMU data and external motion tracking. Future work will address partial leg damage, refining control strategies, and incorporating real-time feedback to further improve adaptability and responsiveness.


\bibliographystyle{IEEEtran}
    \bibliography{IEEEabrv,ref}

\begin{thebibliography}{10}
\providecommand{\url}[1]{#1}
\csname url@samestyle\endcsname
\providecommand{\newblock}{\relax}
\providecommand{\bibinfo}[2]{#2}
\providecommand{\BIBentrySTDinterwordspacing}{\spaceskip=0pt\relax}
\providecommand{\BIBentryALTinterwordstretchfactor}{4}
\providecommand{\BIBentryALTinterwordspacing}{\spaceskip=\fontdimen2\font plus
\BIBentryALTinterwordstretchfactor\fontdimen3\font minus
  \fontdimen4\font\relax}
\providecommand{\BIBforeignlanguage}[2]{{%
\expandafter\ifx\csname l@#1\endcsname\relax
\typeout{** WARNING: IEEEtran.bst: No hyphenation pattern has been}%
\typeout{** loaded for the language `#1'. Using the pattern for}%
\typeout{** the default language instead.}%
\else
\language=\csname l@#1\endcsname
\fi
#2}}
\providecommand{\BIBdecl}{\relax}
\BIBdecl

\bibitem{sun2024fall}
H.~Sun, J.~Yang, Y.~Jia, C.~Zhang, X.~Yu, and C.~Wang, ``Fall prediction,
  control, and recovery of quadruped robots,'' \emph{ISA transactions}, 2024.

\bibitem{li2024dynamic}
S.~Li, Y.~Pang, P.~Bai, S.~Hu, L.~Wang, and G.~Wang, ``Dynamic fall recovery
  control for legged robots via reinforcement learning,'' \emph{Biomimetics},
  vol.~9, no.~4, p. 193, 2024.

\bibitem{lu2024learning}
Y.~Lu, Y.~Dong, J.~Ma, J.~Zhang, and P.~Lu, ``Learning an adaptive fall
  recovery controller for quadrupeds on complex terrains,'' \emph{arXiv
  preprint arXiv:2412.16924}, 2024.

\bibitem{ma2023learning}
Y.~Ma, F.~Farshidian, and M.~Hutter, ``Learning arm-assisted fall damage
  reduction and recovery for legged mobile manipulators,'' in \emph{2023 IEEE
  International Conference on Robotics and Automation (ICRA)}.\hskip 1em plus
  0.5em minus 0.4em\relax IEEE, 2023, pp. 12\,149--12\,155.

\bibitem{zhang2024gait}
Y.~Zhang, H.~Zhang, W.~Ma, P.~Li, and D.~Fang, ``Gait planning and adjustment
  for a hexapod walking robot with one leg failure,'' \emph{Australian Journal
  of Mechanical Engineering}, pp. 1--14, 2024.

\bibitem{ref20}
S.~Terryn, J.~Langenbach, E.~Roels, J.~Brancart, C.~Bakkali-Hassani, Q.-A.
  Poutrel, A.~Georgopoulou, T.~G. Thuruthel, A.~Safaei, P.~Ferrentino
  \emph{et~al.}, ``A review on self-healing polymers for soft robotics,''
  \emph{Materials Today}, vol.~47, pp. 187--205, 2021.

\bibitem{ref21}
M.~Liu, S.~Zhu, Y.~Huang, Z.~Lin, W.~Liu, L.~Yang, and D.~Ge, ``A self-healing
  composite actuator for multifunctional soft robot via photo-welding,''
  \emph{Composites Part B: Engineering}, vol. 214, p. 108748, 2021.

\bibitem{ref22}
K.~Horibe, K.~Walker, and S.~Risi, ``Regenerating soft robots through neural
  cellular automata,'' in \emph{Genetic Programming}, T.~Hu, N.~Louren{\c{c}}o,
  and E.~Medvet, Eds.\hskip 1em plus 0.5em minus 0.4em\relax Cham: Springer
  International Publishing, 2021, pp. 36--50.

\bibitem{ref26}
S.~Reig, E.~J. Carter, T.~Fong, J.~Forlizzi, and A.~Steinfeld, ``Flailing,
  hailing, prevailing: perceptions of multi-robot failure recovery
  strategies,'' in \emph{Proceedings of the 2021 ACM/IEEE International
  Conference on Human-Robot Interaction}, 2021, pp. 158--167.

\bibitem{ozkan2021self}
Y.~Ozkan-Aydin and D.~I. Goldman, ``Self-reconfigurable multilegged robot
  swarms collectively accomplish challenging terradynamic tasks,''
  \emph{Science Robotics}, vol.~6, no.~56, p. eabf1628, 2021.

\bibitem{ref14}
J.~Bongard, V.~Zykov, and H.~Lipson, ``Resilient machines through continuous
  self-modeling,'' \emph{Science}, vol. 314, no. 5802, pp. 1118--1121, 2006.

\bibitem{ref17}
A.~Cully, J.~Clune, D.~Tarapore, and J.-B. Mouret, ``Robots that can adapt like
  animals,'' \emph{Nature}, vol. 521, no. 7553, pp. 503--507, 2015.

\bibitem{cully2015}
A.~Cully, ``Creative adaptation through learning,'' Ph.D. dissertation, Thèse
  de doctorat dirigée par Doncieux, Stéphane et Mouret, Jean-Baptiste
  Informatique Paris 6, 2015.

\bibitem{ref18}
S.~Narain, E.~Mak, D.~Chee, B.~Englot, K.~Pochiraju, N.~K. Jha, and K.~Narayan,
  ``Fast design space exploration of nonlinear systems: Part i,'' \emph{IEEE
  Transactions on Computer-Aided Design of Integrated Circuits and Systems},
  2021.

\bibitem{farid2018fractional}
Y.~Farid, V.~J. Majd, and A.~Ehsani-Seresht, ``Fractional-order active
  fault-tolerant force-position controller design for the legged robots using
  saturated actuator with unknown bias and gain degradation,'' \emph{Mechanical
  Systems and Signal Processing}, vol. 104, pp. 465--486, 2018.

\bibitem{ref27}
K.~Suzuki, H.~Mori, and T.~Ogata, ``Compensation for undefined behaviors during
  robot task execution by switching controllers depending on embedded dynamics
  in rnn,'' \emph{IEEE Robotics and Automation Letters}, vol.~6, no.~2, pp.
  3475--3482, 2021.

\bibitem{feber2022gait}
J.~Feber, R.~Szadkowski, and J.~Faigl, ``Gait adaptation after leg amputation
  of hexapod walking robot without sensory feedback,'' in \emph{International
  Conference on Artificial Neural Networks}.\hskip 1em plus 0.5em minus
  0.4em\relax Springer, 2022, pp. 656--667.

\bibitem{yang2002fault}
J.-M. Yang, ``Fault-tolerant gaits of quadruped robots for locked joint
  failures,'' \emph{IEEE Transactions on Systems, Man, and Cybernetics, Part C
  (Applications and Reviews)}, vol.~32, no.~4, pp. 507--516, 2002.

\bibitem{johnson2010disturbance}
A.~Johnson, G.~Haynes, and D.~Koditschek, ``Disturbance detection,
  identification, and recovery by gait transition in legged robots,'' in
  \emph{2010 IEEE/RSJ International Conference on Intelligent Robots and
  Systems}.\hskip 1em plus 0.5em minus 0.4em\relax IEEE, 2010, pp. 5347--5353.

\bibitem{christensen2014fault}
D.~J. Christensen, J.~C. Larsen, and K.~Stoy, ``Fault-tolerant gait learning
  and morphology optimization of a polymorphic walking robot,'' \emph{Evolving
  Systems}, vol.~5, pp. 21--32, 2014.

\bibitem{pratihar2002optimal}
D.~K. Pratihar, K.~Deb, and A.~Ghosh, ``Optimal path and gait generations
  simultaneously of a six-legged robot using a ga-fuzzy approach,''
  \emph{Robotics and Autonomous Systems}, vol.~41, no.~1, pp. 1--20, 2002.

\bibitem{erden2008free}
M.~S. Erden and K.~Leblebicio{\u{g}}lu, ``Free gait generation with
  reinforcement learning for a six-legged robot,'' \emph{Robotics and
  Autonomous Systems}, vol.~56, no.~3, pp. 199--212, 2008.

\bibitem{shi2022reinforcement}
H.~Shi, B.~Zhou, H.~Zeng, F.~Wang, Y.~Dong, J.~Li, K.~Wang, H.~Tian, and
  M.~Q.-H. Meng, ``Reinforcement learning with evolutionary trajectory
  generator: A general approach for quadrupedal locomotion,'' \emph{IEEE
  Robotics and Automation Letters}, vol.~7, no.~2, pp. 3085--3092, 2022.

\bibitem{zhang2021straight}
F.~Zhang, S.~Zhang, Q.~Wang, Y.~Yang, and B.~Jin, ``Straight gait research of a
  small electric hexapod robot,'' \emph{Applied Sciences}, vol.~11, no.~8, p.
  3714, 2021.

\bibitem{kon2020gait}
J.~Kon and F.~Sahin, ``Gait generation for damaged hexapods using a genetic
  algorithm,'' in \emph{2020 IEEE 15th International Conference of System of
  Systems Engineering (SoSE)}.\hskip 1em plus 0.5em minus 0.4em\relax IEEE,
  2020, pp. 451--456.

\bibitem{bjelonic2021whole}
M.~Bjelonic, R.~Grandia, O.~Harley, C.~Galliard, S.~Zimmermann, and M.~Hutter,
  ``Whole-body mpc and online gait sequence generation for wheeled-legged
  robots,'' in \emph{2021 IEEE/RSJ international conference on intelligent
  robots and systems (IROS)}.\hskip 1em plus 0.5em minus 0.4em\relax IEEE,
  2021, pp. 8388--8395.

\bibitem{future:farghdani_model2}
\BIBentryALTinterwordspacing
S.~Farghdani, O.~Abdelrahman, and R.~Chhabra, ``Fast and modular whole-body
  lagrangian dynamics of legged robots with changing morphology,'' 2025.
  [Online]. Available: \url{https://arxiv.org/abs/2504.16383}
\BIBentrySTDinterwordspacing

\bibitem{farghdani2024singularity}
S.~Farghdani and R.~Chhabra, ``Singularity-free whole-body dynamical equations
  of legged robots for damage simulation,'' in \emph{AIAA SCITECH 2024 Forum},
  2024, p. 1017.

\bibitem{ref16}
M.~S. Erden and K.~Leblebicio{\u{g}}lu, ``Free gait generation with
  reinforcement learning for a six-legged robot,'' \emph{Robotics and
  Autonomous Systems}, vol.~56, no.~3, pp. 199--212, 2008.

\bibitem{porta2004reactive}
J.~M. Porta and E.~Celaya, ``Reactive free-gait generation to follow arbitrary
  trajectories with a hexapod robot,'' \emph{Robotics and Autonomous Systems},
  vol.~47, no.~4, pp. 187--201, 2004.

\bibitem{song1987analytical}
S.-M. Song and K.~J. Waldron, ``An analytical approach for gait study and its
  applications on wave gaits,'' \emph{The International Journal of Robotics
  Research}, vol.~6, no.~2, pp. 60--71, 1987.

\bibitem{suphi2007analysis}
M.~Suphi~Erden and K.~Leblebicio{\u{g}}lu, ``Analysis of wave gaits for energy
  efficiency,'' \emph{Autonomous Robots}, vol.~23, pp. 213--230, 2007.

\bibitem{storn1997differential}
R.~Storn and K.~Price, ``Differential evolution--a simple and efficient
  heuristic for global optimization over continuous spaces,'' \emph{Journal of
  global optimization}, vol.~11, pp. 341--359, 1997.

\bibitem{jethexa}
Hiwonder JetHexa ROS Hexapod Robot Kit Powered by Jetson Nano Boston Dynamics.;
  2022. Available online at: https://www.hiwonder.com/products/jethexa
  (Accessed March, 2024).

\bibitem{8324813}
J.~Sun, J.~Ren, Y.~Jin, B.~Wang, and D.~Chen, ``Hexapod robot kinematics
  modeling and tripod gait design based on the foot end trajectory,'' in
  \emph{2017 IEEE International Conference on Robotics and Biomimetics
  (ROBIO)}, 2017, pp. 2611--2616.

\end{thebibliography}

\end{document}